# Plant Doctor: A hybrid machine learning and image segmentation software to quantify plant damage in video footage

Marc Josep Montagut Marques[1]*, Liu Mingxin[2]*, Kuri Thomas Shiojiri[3]*, Tomika Hagiwara[4], Kayo Hirose[5]**, Kaori Shiojiri[6]**, Shinjiro Umezu[7]**

[1]Department of Integrative Bioengineering, Waseda University, Tokyo, Japan

(E-mail: m.montagut@toki.waseda.jp)

[2]Department of Modern Mechanical Engineering, Waseda University, Tokyo, Japan

(E-mail: verdure.m.liu@ruri.waseda.jp)

[3]Kyoto Prefecture Momoyama High School, Kyoto, Japan

(E-mail: kuri.shiojiri@gmail.com)

[4]Department of Biology, Faculty of Science, Kyushu University, Fukuoka, Japan

(E-mail: tomika.hagiwara@gmail.com)

[5]Department of Anesthesiology and Pain Relief Center, The University of Tokyo Hospital, Tokyo, Japan

(E-mail: hirosek-ane@h.u-tokyo.ac.jp)

[6]Department of Agriculture, Ryukoku University, Otsu, Japan

(E-mail: kaori.shiojiri@agr.ryukoku.ac.jp)

[7]Department of Integrative Bioengineering, Waseda University, Tokyo, Japan

(E-mail: umeshin@waseda.jp)

*First authors

**Corresponding authors

# Abstract

Artificial intelligence has significantly advanced the automation of diagnostic processes, benefiting various fields including agriculture. This study introduces an AI-based system for the automatic diagnosis of urban street plants using video footage obtained with accessible camera devices. The system aims to monitor plant health on a day-to-day basis, aiding in the control of disease spreading in urban areas. By combining two machine vision algorithms, YOLOv8 and DeepSORT, the system efficiently identifies and tracks individual leaves, extracting the optimal images for health analysis. YOLOv8, chosen for its speed and computational efficiency, locates leaves, while DeepSORT ensures robust tracking in complex environments. For detailed health assessment, DeepLabV3Plus, a convolutional neural network, is employed to segment and quantify leaf damage caused by bacteria, pests, and fungi. The hybrid system, named Plant Doctor, has been trained and validated using a diverse dataset including footage from Tokyo's urban plants. The results demonstrate the robustness and accuracy of the system in diagnosing leaf damage, with potential applications in large-scale urban flora illness monitoring. This approach provides a non-invasive, efficient, and scalable solution for urban tree health management, supporting sustainable urban ecosystems.

## Keywords

Urban flora; Plant health monitoring; Quantitative analysis; Deep learning; Hybrid systems

## 1. Introduction

Artificial intelligence has proven to be an advantage towards automatizing diagnostic processes. The agricultural field takes advantage of the computational intelligence of these processes to identify factors that may affect the development of crops. Studies on diagnosing unfavorable conditions such as pest, bacterial, viral or fungal infections have surged in recent years [1-3].

Due to climate change and globalization, local floras are at risk of being exposed to new adverse conditions [4]. Demographic projections point out the acceleration in human population growth [5, 6], stricter policies on nature conservation will arise amidst the necessity to create sustainable futures. Being able to control the traffic of diseases and invasive pests is a priority to protect the ecosystem [7-9]. Advancements in the field of robotics with onboard AIs will support the monitoring of extensive urban and rural areas to immediately deliver disease spread forecasting [10-13].

Trees located in urban areas are beneficial for the development of cities, its conservation results in environment, health and economic advantages [14-18]. Tracking the development of flora consists on the observation of features during long time spans, a task which requires of an exceeding amount of manpower. In this study we developed an AI to automatically diagnose street trees based on videos taken with a commercial camera. The obtained information can be used to monitor tree health on a day-to-day basis.

Our motivation is to control the spread of diseases in urban flora by providing an easy-to-use video analysis software. Video footage of street trees can be obtained from cameras mounted in city maintenance cars. Optimal individual leaf images will be extracted by a machine vision algorithm and analyzed by a convolutional neural network to quantify tree health status. Our non-invasive method will have minimal impact on plant development.

To obtain the best images of each individual leaf we combine two machine vision algorithms, You Only Look Once (YOLO) and DeepSORT. We have chosen YOLOv8 for recognizing the position of the leaves within the frame. YOLOv8 offers the smallest speed to computational power ratio and best stability making it an efficient module to be incorporated within small devices [19], its speed is also beneficial to monitor large populations in real time [20]. We selected DeepSORT to count the recognized leaves. DeepSORT is currently the most advance object tracking algorithm, it can be used in adverse and complex environments making it a robust platform. Giving an individual identification for each leaf is essential for quantifying tree health. Examples of this machine vision network can be found in literature to monitor the growth of crops

[21, 22] and leaf disease detection [23, 24]. Nonetheless, using YOLOv8 to obtain health data from leaves has limited accuracy. The diagnosis for each leaf is binary, rendering it unsuitable for finding features below the threshold. In our study we decided to implement image segmentation to overcome the limitations of YOLOv8.

Image segmentation using convolutional neural networks are commonly used in illness detection in medical applications. In recent studies it was demonstrated that it is a powerful tool to find and quantify diseases on leaves [25-28]. Examples of high contrast diseases such as Rust, Leaf and Slug are demonstrated but also shape diseases such as Leaf Curl can be identified [25]. The versatility of neural networks allow them to be applied in other analytical tasks in the field of agriculture such as counting [29] and morphology reconstruction [30]. However, training data is acquired under optimal lightning and camera angle conditions. For instance, semantic segmentation under complex backgrounds can be affected by environmental factors and disruptions in the shooting. These issues pose significant difficulties for the actual application of semantic segmentation. Our hybrid system software Plant Doctor (PD) combines machine vision and image segmentation machine vision prepares the data under the best conditions to analyze just the Region of Interest (ROI), saving time and processing power.

1. We created a hybrid system combining machine vision and a convolutional neural network to quantify plant damage based on video footage.

2. We trained YOLOv8 to identify individual leaves from plants and extract their ROI and combined it with DeepSORT to track the leaves and find the best frame for data pre-processing.

3. We generated a model by training DeepLabV3Plus using archive images of leaves damaged by bacteria, pests and fungi.

4. We carried out experimentation by obtaining media footage of urban plants and evaluated the system.

5. The results confirm the robustness of the system and demonstrates its use to monitor damage in leaves from a variety of urban flora.

## 2. Materials and Methods

### 2.1 Data Acquisition

The database utilized to train the Plant Doctor model includes footage from common trees and bushes found on the streets of Tokyo, Shinjuku area (Table 1).

A DSLR camera, Nikkon D600 with a 300 mm lens, was used to obtain images and videos. The macro lens is able to obtain close-up details of the leaf texture at longer distances while blurring elements out of the focal plane. Images taken with this method maintain sharpness across the entire frame with minimal aberrations close to the edges.

To expand the database, close-up images of bushes and short trees were taken with a smartphone camera.

Part of the sick leaf data was obtained from Hughes et al. [31] open source repository and our internal database.

The number of images used by each method can be found in Table 2.

## 2.2 Data preprocessing

We prepared the data to facilitate the model training by reducing the frame rate to 3 frames per second and resolution of the videos to 640*640. No additional edition such as color change or filters.

Videos with good lightning condition, in focus and differentiable background were selected for preliminary training. The movement speed of the camera is set to be able to obtain multiple images of the same leave during a one direction pane motion. It was observed that by utilizing a 300 mm lens, out of focus objects are blurred out, facilitating the detections of leaves by the YOLOv8 standard model.

## 2.3 Identifying individual leaves and extracting ROI

The first module was designed to identify and track individual leaves in video footage taken from plants. YOLOv8-PD has been optimized to identify leaves by combining Gold-YOLO [32] and CoTAttention [33].

Gold-YOLO utilizes a novel Gather-and-Distribute mechanism to efficiently exchange information between layers, enabling global fusion of multi-level features, significantly enhancing the model's ability to detect objects with varying sizes. Gold-YOLO achieves a balance between speed and accuracy, making it suitable for real-time object detection tasks, especially for mobile deployment. It surpasses existing YOLO-series models in terms of both accuracy and speed. The GD mechanism of Gold-YOLO can be easily integrated into any existing backbone-neck-head structure, allowing for seamless adoption in various object detection architectures. CoTAttention, proposes a novel neural network architecture for visual recognition tasks. Contextual Transformer Networks (CoTNet) are introduced. CoTNet is an extended Transformer architecture aiming to capture both static and dynamic contexts among input keys. By integrating contextual information through self-attention mechanisms, CoTNet can potentially better understand the relationships between distant parts of an image. By utilizing a 3x3 convolution to contextualize key representations, CoTNet can capture rich contextual information without introducing additional computational overhead. CoTAttention was introduced in the backbone of our model.

The training parameters of the YOLOv8-PD model are set to a batch size of 4, 300 iterations, the chosen optimizer is SGD, at a 0.8 momentum, 0.0003 weight decay with adjusted loss functions at 10 box, 0.7 cls and 1 dfl. We set up data augmentation in the HSV color space as, 0.001 h, 0.5 s, 0.2 v. A comparison between YOLOv8 and YOLOv8-PD can be observed in Fig. 1.

After successfully identifying leaves, DeepSORT is used to track them across frames. DeepSORT is able to handle scenarios where objects may occlude each other temporarily, a common occurrence in the studied footage. Individual images for each frame and leaf can be classified for further analysis into separate stacks (Fig. 2). Laplacian variance [34] and SSIM [35] are then used to select the sharpest image within the stack, we created a score by multiplying each factor, filtering out aberrations caused by motion and ROIs that are too different from each other, such as unframed leaves. This process concludes with the outputting of the best image and its respective DeepSORT identification (Fig. 3).

2.4 Damage quantification.

The second module utilizes DeepLabV3Plus [36] to quantify the amount of damage that has been inflicted to each leaf. DeepLabV3Plus has been chosen due to its advantageous architecture designed for semantic segmentation, it can accurately delineate the boundaries of damaged areas on the leaf surface. The efficient use of training data makes it a suitable choice to generate proof of concept models. It requires fewer annotated samples compared to other methods, facilitating its implementation on diverse sets of plants and trees.

The training process involved manual annotation of a dataset comprising both healthy and unhealthy leaves. Leaf health was characterized by the presence of damage resulting from physical breakage or surface afflictions caused by bacterial or fungal infections. The annotation process, using LabelMe [37] open source software, ensured that the model was trained to accurately discern and classify various types of leaf abnormalities.

To enrich the dataset and enhance model robustness techniques within the Albumentations library [38] were implemented to define a series of image enhancement operations. Horizontal flip, random translation, padding and random cropping, gaussian noise and perspective and random adjustments to contrast, brightness, hue and saturation provided of clearer images to train the model.

The training model used EfficientNet [39] as the encoder. We choose EefficientNet b7 as our pre-trained model. The encoder adopts the Compound Scaling method, which performs coordinated scaling across all dimensions to maximize model performance within a fixed number of computational resources. This capability is particularly beneficial for identifying subtle signs of illness, such as discoloration or texture changes. Additionally, we incorporated the CBAM [40] mechanism to enhance the recognition of small targets.

The activation function was set as a sigmoid to squash the output layer to the binary range. Loss function was set as Binary Cross Entropy Loss to make it compatible with binary classification tasks. The chosen optimizer was Nesterov Accelerated Gradient (NAG), a variant of the standard gradient descent optimization algorithm, a preferred algorithm due to its easy implementation, fast convergence and stability. The resulting model is referred as DeepLabV3Plus-PlantDoctor (DLV3P-PD) and can be seen in Fig. 4.

Model performance was evaluated by a two-step analysis. In the first step DLV3P-PD is used to quantify the total area of the leaf. Subsequently, DLV3P-PD is applied a second time to obtain the damaged area. The first step acts as baseline while the second step is used to calculate the damage ratio. In the last step, the data is saved into a table for further analysis. The data can be evaluated by comparing to a manual annotation as seen in Fig. 5.

The operation flow of the software is explained in Fig. 6.

# 3. Results

## 3.1 Experimental Environment

In this study we utilized Windows 11 operating system with an Intel (R) Core (TM) i3-7100 and and NVIDIA 1070GTX with 8GB of memory. The framework is PyTorch 1.10.1 and Python version is 3.8.

## 3.2 Model training and evaluation

### 3.2.1 Machine vision

For YOLOv8-PD the Stochastic Gradient Descent (SGD) was used as the optimization technique with a batch size of 4 and 300 iterations. The learning rate is 0.01 and 0.937 momentum factor. The model is training using pictures gathered by ourselves using a DSLR camera (22.3%) and smartphone camera (56.7%), additionally we included images from the open-access dataset by PlantVillage (21%) totalling 4595 images.

The performance of YOLOv8-PD was compared to its baseline YOLOv8. The preferred metrics for evaluating YOLO models is mAP@0.5 (mean Average Precision at IoU threshold 0.5), mAP@0.5:0.95 and validation classification loss. The optimized model YOLOv8-PD demonstrated better performance than the baseline. A convergence of 0.99 for the mAP@0.5 is reached at epoch 12 for YOLOv8-PD while YOLOv8 takes 39 epochs. The maximum convergence for YOLOv8-PD for the mAP@0.5:0.95 is 0.954 and YOLOv8 is 0.944. The validation classification loss is smaller for YOLOv8-PD at 0.161 and 0.186 for YOLOv8. The results were experimentally obtained by calculate the average certainty to identify leaves per frame, YOLOv8-PD has a higher detection score average. Comparative data can be observed in Fig. 7.

The model has been used on species with similar morphologic characteristics. The model has been trained to find green oval shaped and ample leaves with minimal clustering. For additional species, a leaf morphology analysis for shape classification such as overall shape, margin type, venation pattern, and other distinct features is required to create tailored machine vision models. Expanding and combining model databases will be beneficial for the automatization of leaf identification in complex scenarios with simultaneous specimens.

In the current state of the software, DeepSORT may encounter problems tracking leaves during frame transition. This problem will cause a double identification of some leaves assigning them multiple IDs. Camera panning direction can also affect tracking, back and forth panning may result in double identification. We envision two approaches to solve this problem. The first solution would be to increase the frame rate of the video for smoother frame transitions. However, a higher frame rate is directly proportional to longer processing times. An alternative solution would be to include 2-dimensional data in the ROI database. Position data in combination with frame count could be used to double check leaves motion progress.

### 3.2.2 Image segmentation

DLV3P-PD training was configured with batch size is set to 2, and the Adam optimizer with a learning rate (lr) of 0.0001 is used here. The training will run for 300 epochs. At the 25th epoch, the learning rate for the decoder will be reduced to 1e-5.

A total of 124 images from our sick leaf database were manually annotated. Images were pre-processed using image enhancement, Gaussian blur with a kernel size of 5x5 is applied to eliminate noise. Then, histogram equalization is performed to enhance image contrast. Finally, the grayscale image is converted to a color image to further enhance natural contrast.

It was found that light reflections on the leaf lead to color change. The green color of the leaves tends to vary to lighter shades when exposed to reflecting light. An additional set of images with reflecting light were added to the training model. As an example, leaf #138 (Fig. 5) had an initial damage ratio of 25.1 %, shiny areas were recognized as damage, with the improved model damage is calculated to 3.81 % a 0.1% difference respect to the manually annotated sample.

DLV3P-PD performance was compared to DeepLabV3Plus and UNet++ (Fig. 8). Overall DLV3P models diagnosed leaf damage more accurately than UNet++. DPLV3P is more suitable for tasks requiring high precision while also being less affected by image variability. This is reflected in the Intersection over Union (IoU) evaluation [41] and Dice Loss [42], also known as Sørensen–Dice coefficient loss [43]. We evaluated (Step 1) Leaf detection and (Step 2) Damage detection separately (Fig. 9). Leaf detection IoU scores are in the excellent range for the DLV3P 0.94 and for the DLV3P-PD 0.96, UNett++ scores 0.82; damage detection scores are in the low range of excellency at 0.74 and 0.76 for DLV3P and DLV3P-PD respectively, UNett++ falls in the very good overlap at 0.68. Dice Loss is larger for UNet++ bot at leaf detection 0.16 and damage detection 0.29, DLV3P and DLV3P-PD have a loss of 0.04, 0.23 and 0.04, 0.21 respectively. We can also observe in the plots that UNet++ is more stable between epochs and a in improvement trend is observed for longer iterations. DLV3P is able to plateau sooner than UNet++.

### 3.3 Plant health diagnosis

The software was used to diagnose a 1 minute video of a tree branch. The total processing time was 50 minutes, with 13 minutes dedicated to extract frame ROI, 16 minutes to copy the highest score ROI for each leaf and 21 minutes to provide damage ratio. The results are shown in a csv file as damage ratios for each of the leaves identified in the video.

Plants with more separate leaves are easier to diagnose as can cause miss recognition in the machine vision module and over leaf area quantification in the segmentation module. Despite of taking videos with a long focal length and creating a depth of field effect the segmentation module can sometimes identify out of focus leaves area and add it to the current subject, thus the preference in more separate leaves.

We used Plant Doctor to analyze footage of four plant species found in Tokyo urban areas. *Ligustrum lucidum, Loropetalum chinense, Litsea japonica* and *Prunus speciosa* were selected due to their similarity to leaves used in the training data but easy to distinguish visual characteristics. Specimens with apparent illness or damage are preferred to test the efficacy of the segmentation module.

We observed that leaf area could be recognized despite of notable shape variations. Details such as veins and brightness did not affect the capability of both the leaf finder module and leaf area segmentation module.

Damage was also recognized despite of variations such as shape and color. The segmentation software is capable of recognizing alterations in the leaf area such as white stains, brown stains, lines and small spots.

The performance of the software compared to manual annotation varies more as damage increases. Heavily damaged leaves may have more shades or variations that affect the capacity of both PD and manual annotation to assess damage quantification. For this reason, it is more difficult to stablish damaged areas from healthy areas and the difference in damage ratio assessment increases.

## 4 Conclusions

By combining YOLOv8 and DeepLabV3Plus a detailed quantification of leaf damage has been extracted from video footage. Characteristics from individual leaves can be analyzed simultaneously from video frames. Video ROI extraction is a powerful tool to create large databases in a short time frame.

The integration of position tracking sensors will give an additional dimension for plant tree health diagnosis. Combining damage and positional data will results in more accurate health assessment for AI models while improving individual leaf tracking.

We expect that this software will help in the ongoing work of botanical experts in the collection of data to improve machine learning models. The development progression is projected towards diagnosing species based on the needs of botanical experts. Their expertise in plant disease will play a key role in the annotation process of damage. In future implementations the differentiation of damage types will add additional diagnosis dimensions, leveraging the effort to individually monitor large individual groups.

# Figures

YOLOv8

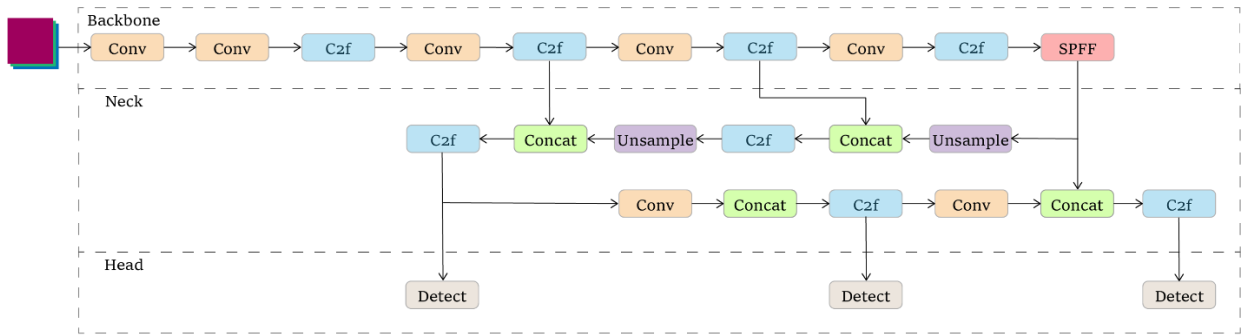

YOLOv8-PD

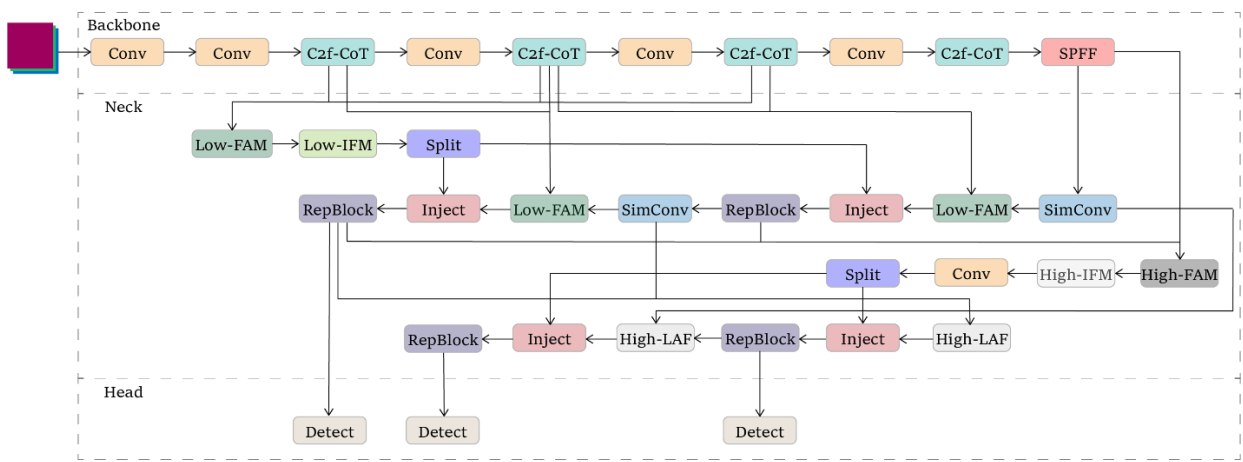

Figure 1 YOLOv8 network structure and YOLOv8-PD structure. Gold-YOLO has been introduced to optimize the detection of objects with varying sizes. CoTAttention has been used to speed up the identification process.

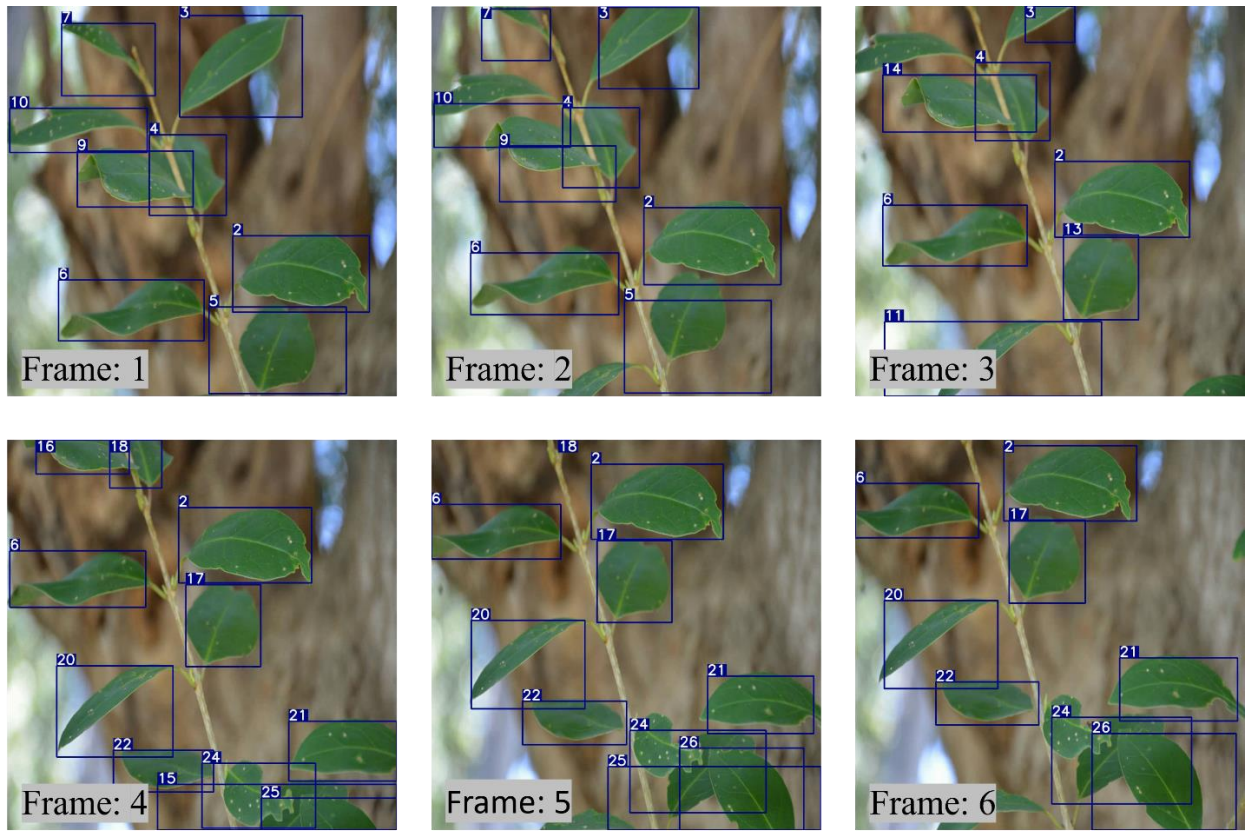

Figure 2 DeepSORT individual leaf tracking. ROI of each leaf is saved with a given identification and frame number. DeepSORT may assign two different IDs to the same leaf if the tracking is lost between frames, a double count is possible, affecting the overall health assesment of the tree.

| ID #5 Frame | ROI | Similarity | Sharpness | Score |
|---|---|---|---|---|
| 1 | 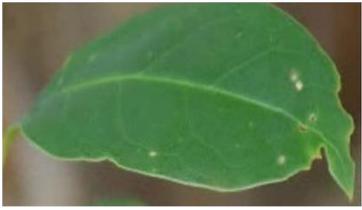 | 0.75 | 94.91 | 71.18 |
| 2 | 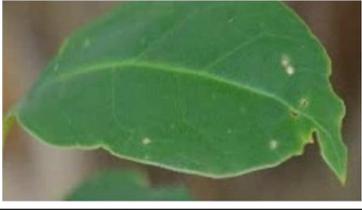 | 0.6 | 101.85 | 61.11 |
| 3 | 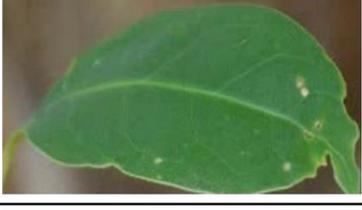 | 0.68 | 121.6 | **82.68** |
| 4 | 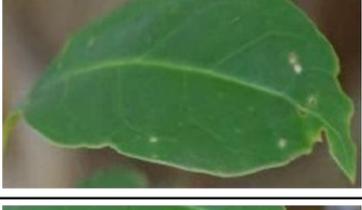 | 0.62 | 69.68 | 43.2 |
| 7 | 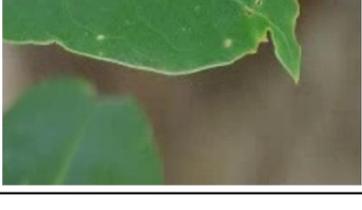 | 0.6 | 101.85 | 61.11 |

Figure 3 Example of a stack obtained from leaf number 5. The score is obtained by multiplying sharpness and similarity values utilizing Laplacian variance and SSIM. In this case frame 3 ROI is selected for further processing due to its higher score.

DeepLabV3Plus-PD

Figure 4 DeepLabV3Plus-PD structure. EfficientNetv3 and CBAM are implemented to optimize leaf damage detection. CBAM is specially effective at recognizing small targets.

| ID | ROI | Step 1: Leaf area | Step2: Damaged area | PD Ratio | Manual Annotation | MA Ratio |
|---|---|---|---|---|---|---|
| #5 | 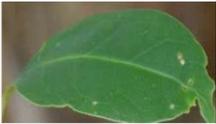 | 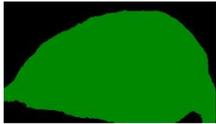 | 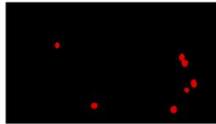 | 1.24 % | 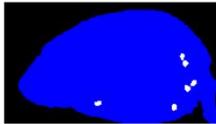 | 1.00 % |
| #21 | 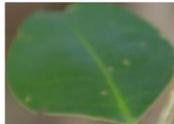 | 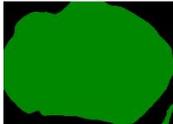 | 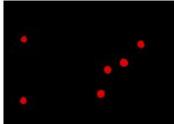 | 1.49 % | 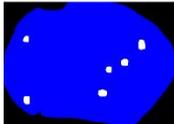 | 1.59 % |
| #69 | 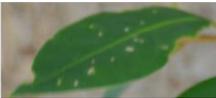 | 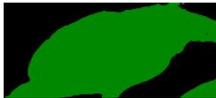 | 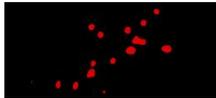 | 4.08 % | 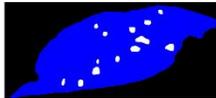 | 4.25 % |
| #138 | 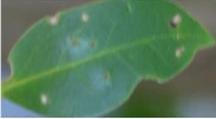 | 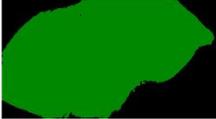 | 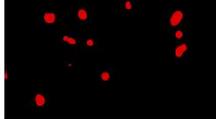 | 3.81 % | 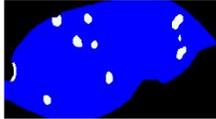 | 3.80 % |
| #143 | 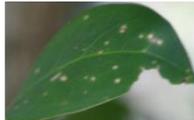 | 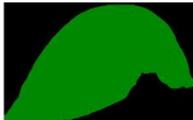 | 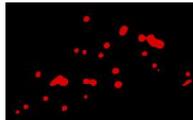 | 6.23 % | 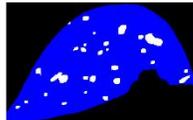 | 5.87 % |

Figure 5 Individual damage ratios for each leaf were calculated by DeepLabV3Plus-DP. The optimized ROIs were afterwards manually annotated using a photography edition software. Manual annotation can be compared to asses the efficacy of the software compared to human observation.

Plant Doctor software

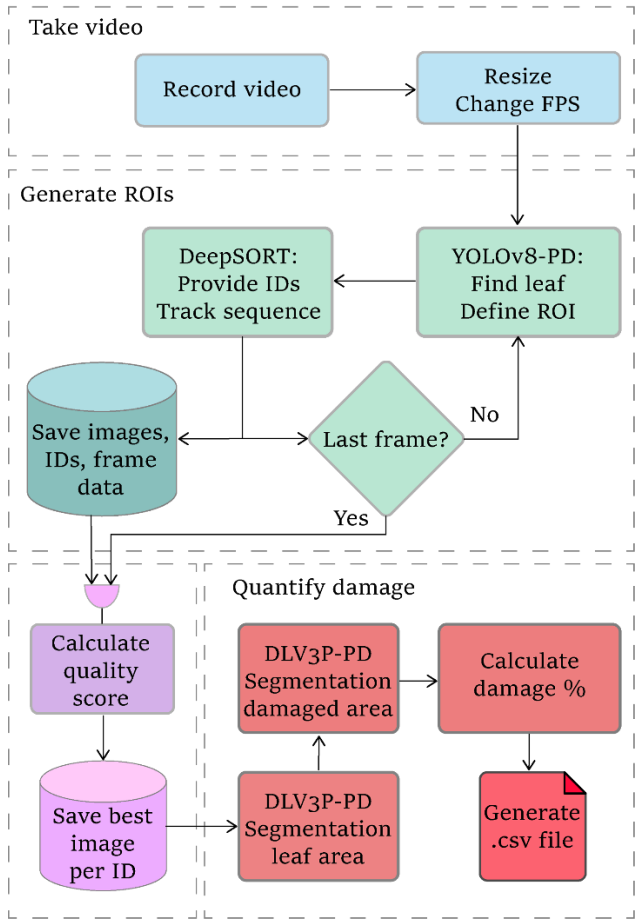

Figure 6 Plant detector software block diagram explaining its work flow. The hybrid system combines machine vision (YOLOv8-PD) and image segmentation (DVL3P-PD) to provide the healthstatus of all individual leaves found in video footage. The produced thada is delivered as a csv file for further statistical analysis. The csv file contains the segmentation result of the best image from each of the leaves, each leaf has an identification number, the frame with best visual characteristics, the area of the leaf, the area of the damage and the ratio of leaf area to damage.

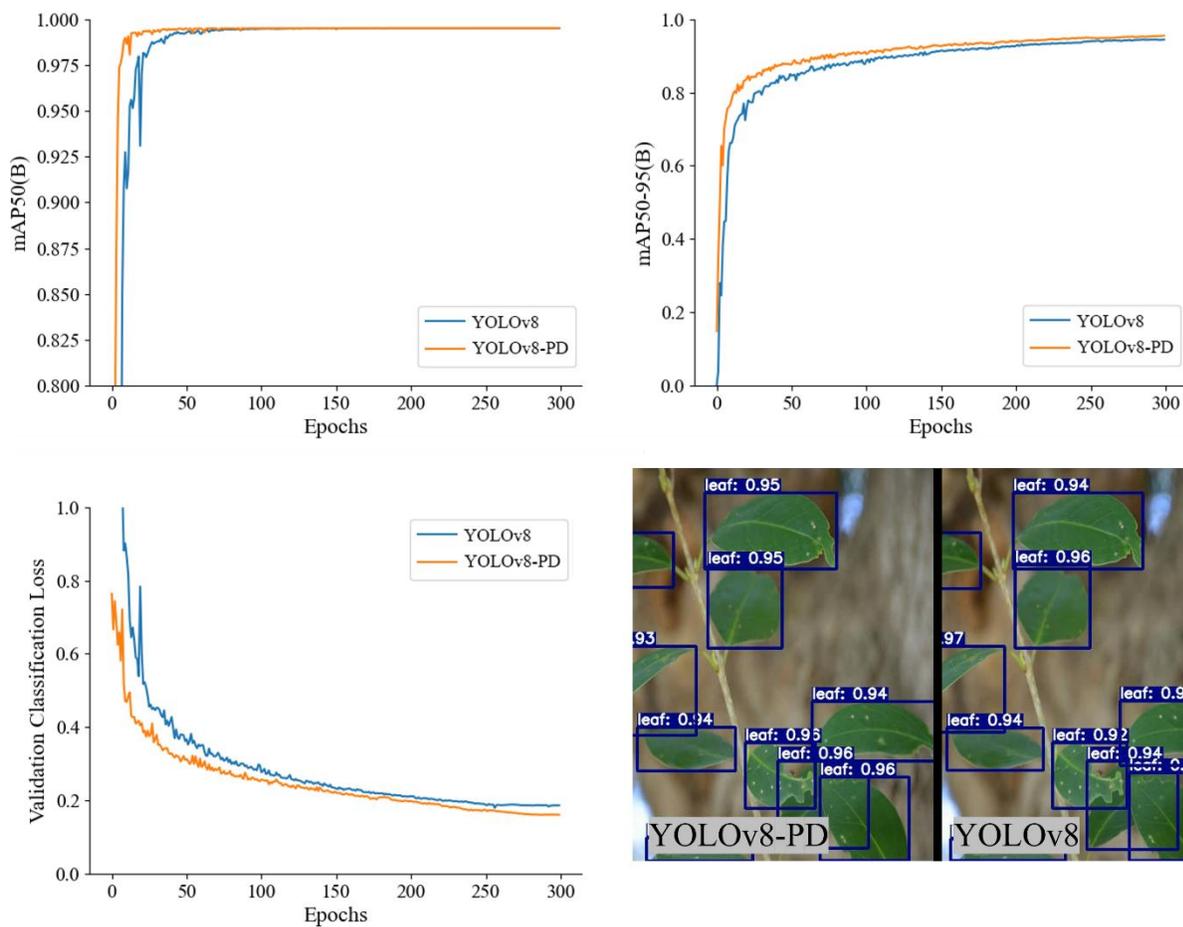

Fig 7 Perfomance comparison of YOLOv8 and YOLOv8-PD. Overall YOLOv8-PD showed a faster stabilization for both mAP50(B) and mAP50-95(B) and fewer loss than YOLOv8. This can be appreciated in the video analysis where the average certainty of found leaves is higher.

| ID | ROI | DLV3P-PD | DeepLabV3Plus | UNet++ | MA Annotation |
|---|---|---|---|---|---|
| #5 | 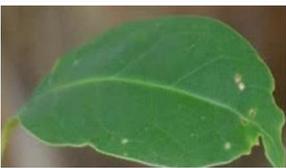 | 1.24 % | 0.56 % | 0.74 % | 1.00 % |
| #21 | 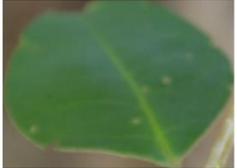 | 1.49 % | 0.98 % | 0.16 % | 1.59 % |
| #69 | 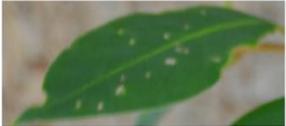 | 4.08 % | 4.1 % | 0.56 % | 4.25 % |
| #138 | 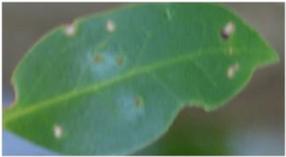 | 3.81 % | 3.00 % | 2.66 % | 3.80 % |
| #143 | 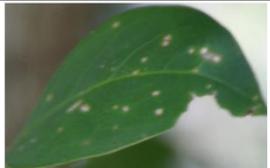 | 6.23 % | 6.54 % | 3.44 % | 5.87 % |

Figure 8 Comparison of the different damage quantification from different image segmentation methods. DLV3P-PD provided more similar results to manual annotation under the same training data. PD variant outperformed its baseline as well as UNet++.

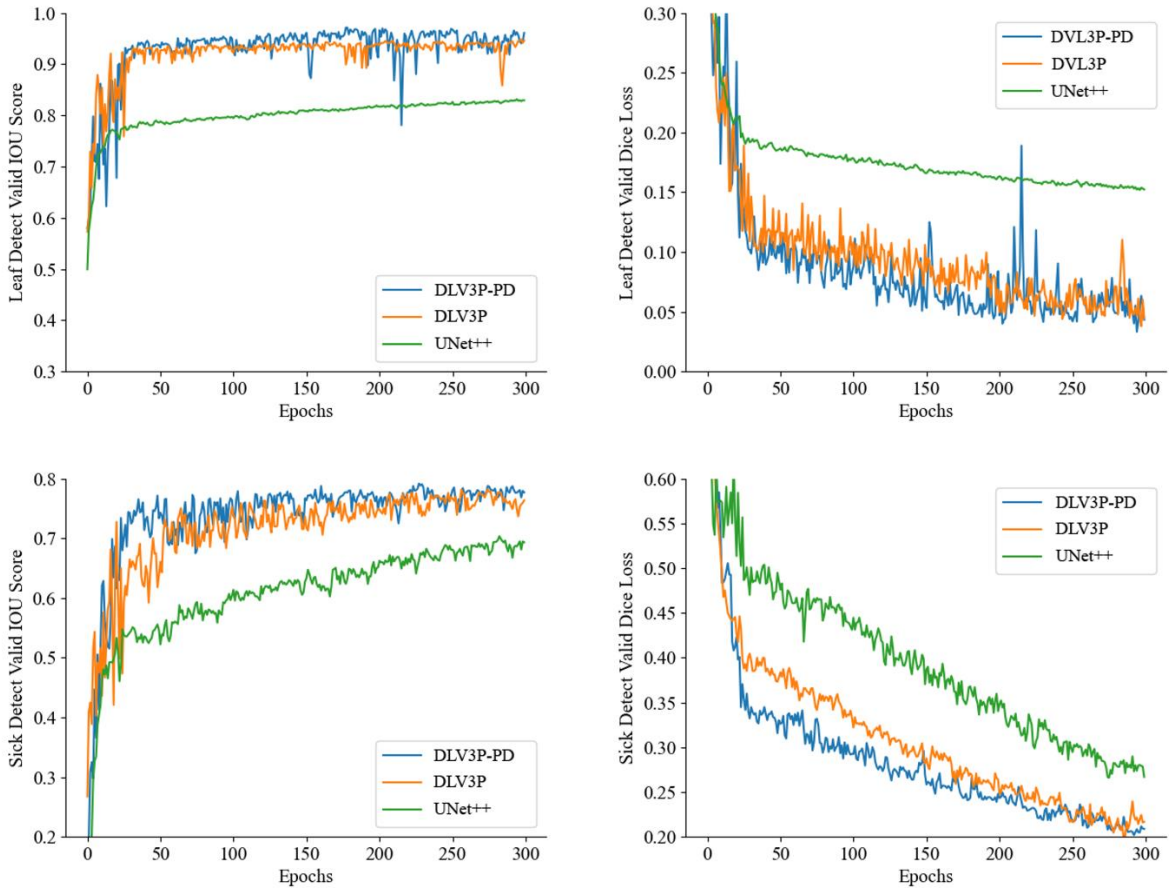

Figure 9 Intersection over Union (IoU) and Dice Loss evaluation for the detection of leaf and damaged areas. DLV3P-PD excels in all metrics for both leaf and damage detection. Leaf detection shows more certainty than damage detection, this could be related to the difficulty of finding damage even in manual annotation.

| Sample frame | Overall appearance | Scientific name | Avg. damaged area per leaf | Avg. difference PD/MA (n=20) |
|---|---|---|---|---|
| 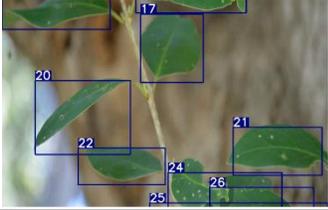 | 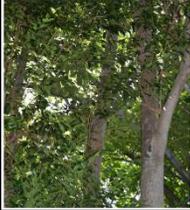 | *Ligustrum lucidum* | 2.90 % | 9.84 % |
| 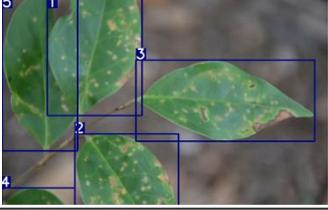 | 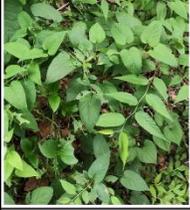 | *Loropetalum chinense* | 5.36 % | 17.97 % |
| 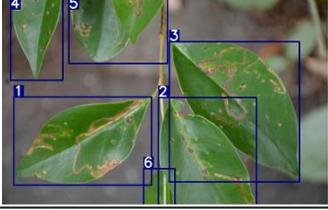 | 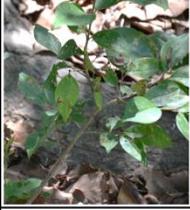 | *Litsea japonica* | 4.45 % | 19.24 % |
| 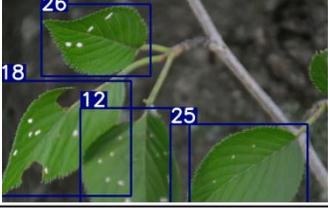 | 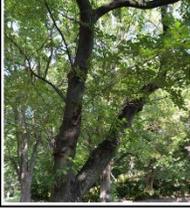 | *Prunus speciosa* | 1.72 % | 11.18 % |

Figure 10 Four different species were analyzed using PD. Leaves with a similar morphology but easy to recognize characteristics were selected. PD was compared to manual annotation to assess the effectivity of the software. Damage assessment presents a higher variation compared to manual annotation when assessing the health of strongly damaged specimens.

## Tables

Table 1

**Trees**

| |
|---|
| *Acer cissifolium* |
| *Cinnamomum yabunikkei* |
| *Cornus kousa* |
| *Lagerstroemia subcostata* |
| *Ligustrum lucidum* |
| *Ligustrum lucidum* |
| *Prunus speciosa* |
| **Bushes** |
| *Camellia japonica* |
| *Cleyera japonica* |
| *Elaegnus pungens* |
| *Hydrangea scandens* |
| *Litsea japonica* |
| *Loropetalum chinense* |
| *Philadelphus satsumi* |

Table 2

| Image type | Quantity | Resolution | Access |
|---|---|---|---|
| DSLR Camera | 1025 | 3936*2624 | 300 mm lens |
| Smartphone | 2608 | 1920*1080 | Close up |
| Sick leaf database | 53 | 1000*1000 | Close up |
| PlantVillage (OA) | 962 | 256*256 | Choose sick |

# Acknowledgements

This study is partially supported by JSPS kaken (23K26069 and 23K26077).